\pdfoutput=1
\documentclass{article}
\usepackage[final]{colm2026_conference}

\usepackage{microtype}
\usepackage{hyperref}
\usepackage{url}
\usepackage{booktabs}
\usepackage{lineno}
\usepackage{graphicx}
\usepackage{amsmath,amssymb}
\usepackage{xcolor}

\definecolor{darkblue}{rgb}{0, 0, 0.5}
\hypersetup{colorlinks=true, citecolor=darkblue, linkcolor=darkblue, urlcolor=darkblue}

\newcommand{\nfacts}{791{,}060}            %
\newcommand{\nprobepoints}{1.58M}
\newcommand{\nprobepointsexact}{1{,}582{,}119}
\newcommand{\ndomains}{13}
\newcommand{\ndeciles}{10}
\newcommand{\nevalsplit}{20{,}003}
\newcommand{\ksamples}{8}
\newcommand{\probetemp}{0.7}
\newcommand{\nmodels}{five}                %

\newcommand{\naliaswiki}{174{,}310}

\newcommand{\llamaunkar}{93.0\%}

\newcommand{\gemmaknownar}{6.5\%}
\newcommand{\gemmaknownen}{10.0\%}

\newcommand{\qwenunkar}{74.8\%}

\newcommand{\gptossunkar}{75.2\%}
\newcommand{\gptossunken}{67.2\%}
\newcommand{\unkrangeall}{67--93\%}        %

\newcommand{\unkjacrange}{0.75--0.91}      %
\newcommand{\knownjacrange}{0.16--0.61}    %
\newcommand{\withinfamknownar}{0.40}       %
\newcommand{\withinfamknownen}{0.50}
\newcommand{\nestingprob}{0.98}            %
\newcommand{\tailunknownpct}{80\%}         %
\newcommand{\topdecileunkmin}{58\%}        %

\newcommand{\unkjacexcess}{0.07--0.29}     %
\newcommand{\knownliftrange}{5--12$\times$}%
\newcommand{\unkliftrange}{1.1--1.6$\times$}%
\newcommand{\retestknownjacceil}{0.87}     %
\newcommand{\retestunkjacceil}{0.98}       %
\newcommand{\transferprecrange}{0.76--0.99}%
\newcommand{\transferrecrange}{0.79--0.99} %
\newcommand{\transferliftrange}{4--22}     %

\newcommand{\scaleknownarfrom}{6.5\%}\newcommand{\scaleknownarto}{12.9\%}
\newcommand{\scaleknownenfrom}{10.0\%}\newcommand{\scaleknownento}{17.0\%}
\newcommand{\scaleunkarto}{83.1\%}
\newcommand{\scaleunkento}{80.0\%}

\newcommand{\popunkrangegemma}{0.91$\rightarrow$0.82}  %

\newcommand{\thinkknownar}{9.0\%}\newcommand{\thinkunkar}{78.9\%}\newcommand{\thinkinvar}{7.2\%}
\newcommand{\thinkknownen}{11.7\%}\newcommand{\thinkunken}{75.2\%}\newcommand{\thinkinven}{8.0\%}
\newcommand{\thinkpairunkjacar}{0.852}     %

\newcommand{\elicrange}{91--98\%}          %

\newcommand{\elicqwenar}{91.2\%}

\newcommand{\hedgellamaar}{51.5\%}\newcommand{\hedgellamaen}{5.0\%}
\newcommand{\hedgegemmaar}{5.5\%}\newcommand{\hedgegemmaen}{0.1\%}
\newcommand{\hedgegemmabigar}{4.7\%}\newcommand{\hedgegemmabigen}{0.1\%}
\newcommand{\hedgeqwenar}{70.1\%}\newcommand{\hedgeqwenen}{21.4\%}
\newcommand{\hedgegptossar}{15.7\%}\newcommand{\hedgegptossen}{9.6\%}

\newcommand{\konecondrange}{0.57--0.79}    %
\newcommand{\ktwocondrange}{0.75--0.87}
\newcommand{\kfourcondrange}{0.88--0.94}

\newcommand{\tiersavings}{2--3$\times$}
\newcommand{\koneprecrange}{0.91--0.97}    %
\newcommand{\konebaserange}{0.71--0.88}    %
\newcommand{\koneprecliftmin}{8}           %
\newcommand{\koneprecliftmax}{20}          %

\newcommand{\latentaucrange}{0.89--0.95}
\newcommand{\latentauclama}{0.953}         %
\newcommand{\latentaucgemma}{0.923}        %
\newcommand{\latentaucgemmabig}{0.890}     %
\newcommand{\latentlmatchedbig}{0.831}
\newcommand{\latentdepthpct}{50--69\%}     %

\newcommand{\hedgegoldmax}{0.21\%}         %
\newcommand{\hedgegoldgemma}{0.004\%}

\newcommand{\retestunknown}{98.9\%}\newcommand{\retestknown}{93.3\%}
\newcommand{\retestweak}{59.6\%}\newcommand{\retestbalanced}{84.7\%}

\newcommand{\graderrllama}{4.6\%}\newcommand{\graderrgemma}{5.8\%}
\newcommand{\graderrqwen}{10.8\%}\newcommand{\graderrgptoss}{13.4\%}
\newcommand{\graderrrange}{4.6--13.4\%}
\newcommand{\aliaspiloteffect}{6$\rightarrow$30/200}

\newcommand{\judgekapparange}{0.89--0.92}  %
\newcommand{\judgeagreerange}{94.5--95.9\%}

\newcommand{\goldauditn}{1{,}500}
\newcommand{\goldauditbad}{0}
\newcommand{\goldauditbadub}{0.2\%}        %
\newcommand{\goldauditinsuff}{12.1\%}
\newcommand{\goldsuffrate}{87.9\%}         %

\newcommand{\routerhallred}{56$\times$}    %

\newcommand{\rfourn}{6{,}000}
\newcommand{\rfourstillinvalid}{50.7\%}
\newcommand{\rfourtounknown}{45.2\%}
\newcommand{\rfourtoknown}{1.3\%}

\newcommand{\latentpoponly}{0.60}          %

\newcommand{\wasteunkmin}{98.8\%}          %
\newcommand{\wasteunkgemma}{99.6\%}\newcommand{\wasteunkqwen}{98.9\%}\newcommand{\wasteunkgptoss}{99.1\%}
\newcommand{\wasteknownrange}{6--18\%}     %
\newcommand{\wasteoverallgemma}{93\%}\newcommand{\wasteoverallqwen}{79\%}
\newcommand{\thinkunkgemma}{1{,}062}\newcommand{\thinkknowngemmatok}{561}  %
\newcommand{\thinkunkgptoss}{412}\newcommand{\thinkknowngptoss}{210}
\newcommand{\thinktrunc}{5.9\%}            %
\newcommand{\thinktruncunk}{6.5\%}         %

\newcommand{\neffortsubset}{150{,}001}
\newcommand{\effortlowthink}{73}\newcommand{\effortmedthink}{359}\newcommand{\efforthighthink}{1{,}420}
\newcommand{\effortlowknownar}{11.5\%}\newcommand{\effortmedknownar}{14.6\%}\newcommand{\efforthighknownar}{16.1\%}
\newcommand{\effortlowknownen}{21.2\%}\newcommand{\effortmedknownen}{23.2\%}
\newcommand{\effortwaste}{99.1\%}          %
\newcommand{\efforthighwaste}{99.6\%}      %
\newcommand{\efforthighinvalidar}{16.4\%}  %
\newcommand{\efforthighinvaliden}{15.7\%}

\newif\iffinalnumbers
\finalnumberstrue
\iffinalnumbers
  
\else
  
\fi

\title{Ignorance Is Shared, Knowledge Is Private: \\ Efficiently Mapping Knowledge Boundaries}

\author{Owais Aijaz, Nikhil Ranjan, Maaz Maqsood, Ahmer Tabassum, Haonan Li, \\
\bf Hector Xuguang Ren, Zhengzhong Liu, Preslav Nakov, Eric P. Xing \\[3pt]
Institute of Foundation Models, MBZUAI \\
\texttt{owais.aijaz@mbzuai.ac.ae}}

\begin{document}

\ifcolmsubmission
\linenumbers
\fi

\maketitle

\lhead{Published at the 2nd Workshop on Efficient Reasoning, COLM 2026}

\begin{abstract}
Large language models hallucinate most on facts that are weakly represented in
their training data, and standard remedies (sampling more answers, or
generating longer reasoning traces) spend compute without first asking
whether the model knows the answer at all. We study this question on
\nfacts{} evidence-grounded Arabic/English long-tail facts, giving
\nprobepoints{} behavioral probe points per model across \nmodels{} open
LLMs (8B--120B). The long tail is closed to everyone: every model leaves
\unkrangeall{} of the bank unknown per language, and unknown sets overlap
across model families with Jaccard similarity \unkjacrange{}, above the overlap their sizes alone would produce in every pair. As a result, one model's unknown map flags another's unknowns at \transferprecrange{} precision without probing the second model. Known sets, in contrast, overlap far less, and scaling
within a family leaves the tail largely unknown. This structure makes
boundary detection cheap exactly where it matters: a single sampled answer flags unknown facts at \koneprecrange{} precision, a linear probe over hidden states separates known from unknown with no generation at AUC \latentaucrange{}, and only the rarer known-vs-weak distinction needs $k\ge4$ samples. A tiered budget therefore cuts labeling cost \tiersavings{} versus uniform sampling. Reasoning, in contrast, buys little knowledge at high cost: models think up to twice as hard on facts they do not know, the effort dial changes how long a model spends before a wrong answer rather than whether the answer is wrong, and $\ge$\wasteunkmin{} of reasoning tokens spent on unknown facts precede a wrong answer in every model. Before spending tokens thinking, a system should first estimate whether it is inside its knowledge boundary.
\end{abstract}

\section{Introduction}
\label{sec:intro}

If a language model never learned a fact, no amount of inference-time
compute will produce it. Yet the standard inference-time remedies for
hallucination, such as self-consistency sampling and longer chains of
thought, spend tokens before asking a more basic question: does the model
know this at all? For most of the world's culturally local knowledge, the
answer is no. We use Arab-culture facts from Arabic Wikipedia as a stress
case. These facts are evidence-grounded, verifiable, and routinely
ingested by open pretraining recipes, yet, as we show, nearly all of them
fall outside every open model's knowledge boundary.

This paper maps that boundary at scale and asks what the map is worth to
a practitioner who cares about efficiency. We probe \nmodels{} open LLMs
(Llama-3.1-8B, \citealp{llama3herd2024}; Gemma-4-12B and 31B,
\citealp{gemma4report2026}; Qwen3.6-35B, \citealp{qwen36blog2026};
gpt-oss-120B, \citealp{gptoss2025}; exact variants:
Table~\ref{tab:models}) on
\nfacts{} bilingual evidence-grounded facts, which gives \nprobepoints{}
(fact, language) probe points per model with \ksamples{} samples each,
and label every point known, weak, or unknown following the sampling
protocol of \citet{gekhman2024finetuning}.

We report three main findings.

\textbf{Unknown sets are shared across models; known sets are not
(\S\ref{sec:boundaries}).}
Unknown sets overlap across model \emph{families} at Jaccard
\unkjacrange{}, near the same-model retest ceiling, while known sets
overlap at only \knownjacrange{}. The absolute asymmetry is what matters
for reuse, and we test reuse directly: one model's unknown set flags
another's at \transferprecrange{} precision. Scaling within a family
(Gemma-4 12B$\to$31B) nearly doubles known mass yet leaves
$\ge$\tailunknownpct{} of the long tail unknown, and knowledge nests
upward: what the 31B does not know, the 12B almost surely does not know
either. Prior work showed that \emph{errors} correlate across models
\citep{kim2025correlated} and that the long tail resists scaling
\citep{kandpal2023longtail}; the boundary-labeled view shows the shared
data gap directly, per fact.

\textbf{The boundary is cheap to find, and the cost is asymmetric
(\S\ref{sec:cost}).} Unknowns dominate the bank and are stable under
sampling, so a single sample flags them at \koneprecrange{} precision,
and a linear probe on hidden states, following
\citet{gottesman2024estimating}, separates known from unknown at AUC
\latentaucrange{} with zero generated tokens. The expensive distinction
is known-vs-weak, which needs $k\ge4$ samples but applies only to the
small non-unknown slice. Our contribution is the cost/fidelity curve of
the labeling protocol, not the probe: a tiered budget cuts labeling cost
\tiersavings{} versus uniform $k=\ksamples$ sampling at matched
fidelity.

\textbf{Reasoning adds little recall and costs many tokens
(\S\ref{sec:reasoning}).} With full-bank think-on/think-off contrasts
and a three-level effort dial on gpt-oss-120B, we track where reasoning
tokens go, conditioned on boundary labels. Models spend up to 2$\times$
more thinking on unknowns than on knowns, so compute escalates exactly
where it cannot help; thinking lifts known mass by only a few points
while the Arabic tail stays largely unknown; and on unknown facts,
$\ge$\wasteunkmin{} of reasoning tokens precede a wrong answer in every
model. A router simulated from the same samples (\S\ref{sec:router})
trades $\sim$1.6pp accuracy for a \routerhallred{} hallucination
reduction at zero reasoning tokens. Concurrent work reports aggregate
accuracy or abstention shifts from thinking
\citep{zhao2025tts,thinking2recall2026}; conditioning on boundary labels
turns those aggregates into per-token accounting of what reasoning
actually retrieves.

\paragraph{Contributions.}
(C1) To our knowledge, the largest multi-model knowledge-boundary map to
date, and the first at any scale for a culturally long-tail Arabic
corpus: \nprobepoints{} labeled probe points per model across
\nmodels{} open models in two languages, with null-model-calibrated
overlap structure, upward nesting, and a directly tested cross-model map
transfer (\S\ref{sec:boundaries}).
(C2) The first cost/fidelity accounting of boundary \emph{labeling}
itself: the asymmetric $k$-sample curve with the established latent
probe as its $k=0$ point, yielding a tiered budget \tiersavings{}
cheaper at matched fidelity, in GPU-hours (\S\ref{sec:cost}).
(C3) A full-scale accounting of when reasoning tokens buy recall,
conditioned on boundary labels, across think toggles and a three-level
effort dial, closed by a sample-level routing simulation
(\S\ref{sec:reasoning}).
All probing, grading, and analysis artifacts will be released.

\section{Related work}
\label{sec:related}

We summarize the closest lines here; Appendix~\ref{app:related} gives a
fuller treatment. \citet{gekhman2024finetuning} introduced the
sampling-based known/weak/unknown taxonomy (SliCK) and showed that
fine-tuning on unknown facts teaches hallucination; abstention SFT and
boundary-aware training build on such probes
\citep{zhang2024rtuning,yang2025barrel,gao2026karl,cheng2024idk}, and
\citet{li2024kbsurvey} survey the area. Errors are known to correlate
across models \citep{kim2025correlated,goel2025great} and the long tail
resists scaling
\citep{kandpal2023longtail,mallen2023popqa,sun2024headtotail}; we instead
measure boundary-labeled \emph{unknown sets} on a fixed evidence-grounded
corpus, where the overlap asymmetry and upward nesting become visible.
Pre-generation probes on hidden states are an established alternative to
sampling
\citep{gottesman2024estimating,kossen2024sep,orgad2025llms,xiao2025crosslingual};
we claim no novelty for the probe and contribute the cost/fidelity curve
of the labeling protocol itself. Adaptive-RAG routes between parametric
answering and retrieval on self-knowledge or uncertainty signals
\citep{wang2023skr,jeong2024adaptiverag,moskvoretskii2025adaptive}; our
recipe supplies the boundary-side accounting that this mechanism needs.
On reasoning vs.\ knowledge, test-time scaling does not help
knowledge-intensive tasks \citep{zhao2025tts}, thinking expands recall
only modestly \citep{thinking2recall2026}, and reasoning models fail on
unanswerable questions \citep{kirichenko2025abstentionbench}; we
condition token expenditure on independently measured boundary labels at
791K-fact scale. To our knowledge no Arabic (or comparably culturally
long-tail) knowledge-boundary map exists at any scale
\citep{arahallueval2025,multiloko2025}.

\section{A culturally long-tail fact bank and probe protocol}
\label{sec:methods}

\subsection{Fact bank}
\label{sec:factbank}
We build \nfacts{} evidence-grounded QA facts from the Arabic Wikipedia
2025-07 snapshot using a generator model (GLM-5.2,
\citealp{glm5report2026}) that is \emph{excluded from all evaluation} in
this paper. Each fact includes a native-Arabic question and a
native-English question (independently authored, not translated), one or
more gold answers, a verbatim evidence sentence from the source article,
an entity domain, and a popularity score from pageviews. The bank is
stratified over \ndomains{} domains $\times$ \ndeciles{} popularity
deciles. An LLM judge verified every fact against its evidence span; in
a judge bake-off, agreement was 92.8\% ($\kappa=0.616$), and we adopted
the stricter judge. Before any downstream use, we froze an evaluation
split of \nevalsplit{} facts, sampled proportionally over
domain$\times$decile cells. These facts are probed and their labels
appear in the aggregate maps (the \nprobepointsexact{} probe points of
Figure~\ref{fig:f1} cover the full bank), but the split itself is
reserved for end-to-end evaluation of trained or routed systems, which
no experiment in this paper performs.

\subsection{Behavioral boundary probe}
\label{sec:probe}
Following \citet{gekhman2024finetuning}, we draw $k=\ksamples$ samples at
$T=\probetemp$ per (fact, language, model) through each model's own chat
template, in its \emph{deploy-default} reasoning mode (Gemma-4: no-think;
Qwen3.6 and gpt-oss: reasoning; Llama-3.1: non-reasoning). Reasoning and
answer channels are recorded separately with per-family parsing
(Appendix~\ref{app:probe}). A sample is valid if generation stopped naturally or
produced a non-empty answer; records with $<$50\% valid samples are labeled
invalid and excluded (per-model invalid and truncation rates:
Appendix~\ref{app:probe}). Labels are assigned over valid samples, in
order of precedence: \textbf{known} if $\ge$75\% are correct, else
\textbf{weak} if $\ge$25\% are correct, else \textbf{unknown}. Unlike
SliCK we do not use a separate greedy-decoding gate; \S\ref{sec:cost}
shows label stability under subsampling, which subsumes the greedy
question in practice.

\subsection{Grading: evidence-grounded, alias-robust}
\label{sec:grading}
Correctness is normalized substring match (Arabic diacritic stripping,
alef/yaa/taa unification, digit folding, word-boundary rules for short
numeric golds, componentwise dates), with two guards against systematic
false positives: an answer that hedges is never correct, and golds
contained verbatim in the question never match. An exhaustive audit
bounds hedged-but-correct relabeling at \hedgegoldmax{} of
unknown-labeled facts in the heaviest-hedging cell
($\le$\hedgegoldgemma{} elsewhere), so the unknown mass is not a hedging
artifact. Because native-Arabic golds alone would mislabel many English
answers as unknown (in a pilot, English known counts rose
\aliaspiloteffect{} once aliases were added), we grade against a union of
gold aliases from Arabic-Wikipedia langlinks and redirects (\naliaswiki{}
facts) and, for the non-entity tail, renderings generated at $T=0$ by
GLM-5.2 (never probed, and forbidden from answering the question). Alias
provenance, retirement policy, and homograph bounds:
Appendix~\ref{app:probe}.

\paragraph{Grader validation.}
Two semantic judges, independent of gold authorship and of the map they
grade, re-grade $\ge$1{,}000 stratified samples per map
(Appendix~\ref{app:grader}). Population-weighted grader error is
\graderrllama{}--\graderrgptoss{} (Table~\ref{tab:graderval}) and
false-negative-dominant in every map: the semantic judge credits answers
the substring grader rejects, so known masses are, if anything,
understated. Inter-judge agreement is \judgeagreerange{} ($\kappa$
\judgekapparange), and an evidence-anchored audit of \goldauditn{}
stratified golds finds \goldauditbad{} contradicted (95\% upper bound
\goldauditbadub).

\subsection{Latent probe}
\label{sec:latentmethod}
For each dense model we take 5{,}000 facts balanced over
$\{$ar,en$\}\times\{$known, unknown$\}$ (unknowns restricted to
$0/\ksamples$ correct), run one forward pass per question (chat template,
pre-first-answer-token state), and fit a per-layer linear classifier (exact
ridge in Gram space, 5-fold CV) on the last prompt token's hidden state.
Controls against surface artifacts (layer-0, shuffled labels, constant
feature, length-only, popularity-only, length-matched resampling):
Appendix~\ref{app:latent}.

\subsection{Cost accounting}
\label{sec:costmethod}
Every pass logs H100 node-hours ($\times$8 GPUs/node). $k$-fidelity
recomputes labels from random $k$-subsamples ($k\in\{1,2,4\}$, 5 repetitions)
against the $k=\ksamples$ reference, both unconditionally and conditioned on
non-unknown reference labels; the latent probe supplies the $k=0$ point
(Table~\ref{tab:ledger}).

\section{Boundary maps: ignorance is shared, knowledge is private}
\label{sec:boundaries}

\begin{figure}[t]
\centering
\includegraphics[width=\linewidth]{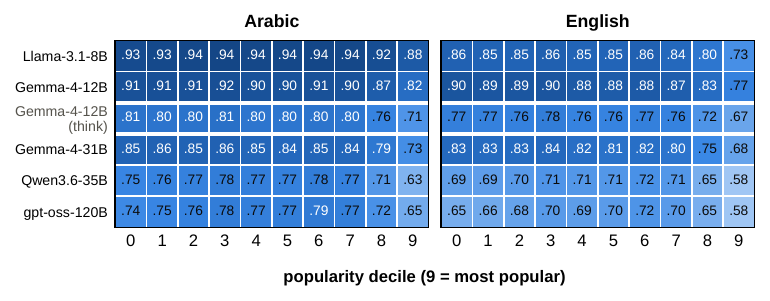}
\caption{Unknown rate by model, popularity decile, and language
(\nprobepointsexact{} labeled probe points per map); the color scale spans
0.5--1.0 since no cell falls below it. Even the most popular decile stays
majority-unknown for every model.}
\label{fig:f1}
\end{figure}

\begin{figure}[t]
\centering
\includegraphics[width=\linewidth]{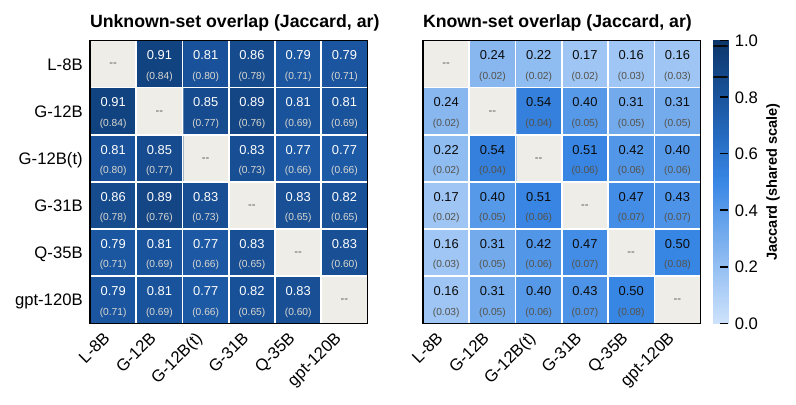}
\caption{Pairwise Jaccard overlap of unknown (left) vs.\ known (right)
sets, Arabic; shared 0--1 color scale (English: Appendix~\ref{app:maps},
same pattern). Small gray values give the per-pair expected Jaccard under
independence given the set masses: unknown overlap sits \unkjacexcess{}
above chance, known overlap \knownliftrange{} above it. Ticks on the
color bar mark the same-model fresh-draw retest ceilings
(\retestunkjacceil{} unknown, \retestknownjacceil{} known;
\S\ref{sec:limitations}): cross-family unknown overlap runs close to the
most that a re-measurement of the \emph{same} model attains.}
\label{fig:f2}
\end{figure}

\paragraph{Most of the bank is unknown to every model.}
Unknown mass ranges \unkrangeall{} per language across the five
deploy-default maps (Llama-3.1-8B \llamaunkar{} ar to gpt-oss-120B
\gptossunken{} en; Figure~\ref{fig:f1}). Even the \emph{most popular}
decile never drops below \topdecileunkmin{} unknown for any model
(\popunkrangegemma{} across deciles, Gemma-4-12B ar). Popularity barely
helps, and English fares only modestly better.

\paragraph{Unknown sets overlap across families far more than known sets.}
Across the nine cross-family deploy-default pairs, unknown-set Jaccard
overlap is \unkjacrange{} while known-set overlap is \knownjacrange{}, and
unknown overlap exceeds known overlap in \emph{every} pair
(Figure~\ref{fig:f2}). The same holds once within-family and
matched-think pairs are included, in all 15 pairs among our six maps.

Raw Jaccard between sets of unequal mass must be read against a null
model; Figure~\ref{fig:f2} reports the per-pair independence
expectations alongside the observed values. Relative to chance,
\emph{both} sides are structured, and knowledge the more strongly
(\knownliftrange{} its null vs.\ \unkliftrange{} for ignorance), but the
raw asymmetry is a statement about \emph{absolute} agreement, which is
what map reuse cares about. A same-model fresh-draw retest
(\S\ref{sec:limitations}) caps how much overlap any measurement can
attain: cross-family unknown overlap runs near that ceiling, about as
tight as measurement noise allows, while known overlap sits far below
its own. Shared ignorance at near-ceiling overlap is what a shared
training-data coverage gap would produce: what the common web-derived
distribution lacks is absent for everyone, while what \emph{is} learned
from sparse exposures differs from run to run (we observe the behavioral
signature, not the corpora).

\paragraph{One model's map predicts another's unknowns.}
Using model $A$'s unknown set to flag model $B$'s unknowns achieves
precision \transferprecrange{} at coverage \transferrecrange{} across all
18 directed cross-family pairs, in both languages. This beats $B$'s
marginal unknown rate, the constant-predictor baseline, by
\transferliftrange{}pp in every direction, with the largest lifts
between the two reasoning models. The lift is modest only because the base rate is extreme: a new model's map starts \transferprecrange{} correct before a single
probe is drawn.

\paragraph{Scaling adds known facts but leaves the tail unknown.}
Within the Gemma-4 family, 2.6$\times$ more parameters nearly doubles known
mass (Arabic \scaleknownarfrom$\to$\scaleknownarto, English
\scaleknownenfrom$\to$\scaleknownento) yet unknown mass falls only to
\scaleunkarto/\scaleunkento{} (ar/en); the tail stays
$\ge$\tailunknownpct{} unknown. Knowledge nests upward:
$p(\text{12B unknown} \mid \text{31B unknown}) = \nestingprob$;
within-family known overlap (\withinfamknownar{} ar, \withinfamknownen{} en)
exceeds cross-family overlap but remains far from identity. The pattern
extends to the largest models: gpt-oss-120B and Qwen3.6-35B, both
reasoning-mode, still leave \gptossunkar/\qwenunkar{} of Arabic facts
unknown. Scaling is an inefficient way to buy the long tail.

\paragraph{Unknowns are missing knowledge, not failed extraction.}
A natural objection: Wikipedia is a standard open-pretraining ingredient,
so ``unknowns'' might be known but unsurfaceable under this prompt format.
Two measurements say otherwise. First, elicitability: \elicrange{} of
unknown-labeled facts never produce the gold in any of \ksamples{} samples
(96--98\% non-reasoning; 91--92\% reasoning, whose thinking occasionally
surfaces a gold it then abandons). There is no sizable
knows-but-won't-say reservoir, and hedged answers containing the gold are
$\le$\hedgegoldmax{} of unknowns (\S\ref{sec:grading}). Second, the
latent probe (\S\ref{sec:cost}) detects these facts as unknown from the
model's own hidden state: the internal state distinguishes them, yet the
gold is never emitted. Whether storage gap or irrecoverable extraction,
it is real for any practical purpose.

\paragraph{Hedging rates depend on language.}
Models hedge (``I don't know'') on their unknowns at rates that differ
sharply \emph{by language within the same model} (Llama-3.1: \hedgellamaar{}
of Arabic unknowns vs.\ \hedgellamaen{} of English ones), and the
Arabic$>$English gap replicates across all five models with model-specific
magnitude (per-model rates: Appendix~\ref{app:maps}), consistent with
multilingual abstention deficits under fine-tuning
\citep{feng2024abstainlang}. Abstention is neither a stable model property
nor transferable across languages, which is another argument for external
boundary detection.
\section{How cheaply can you find the boundary?}
\label{sec:cost}

Boundary-conditioned pipelines \citep{gekhman2024finetuning,zhang2024rtuning}
implicitly assume the map is affordable. It is not free: at $k=\ksamples$
samples per (fact, language), all experiments in this paper total
1{,}243 H100 node-hours ($\approx$9{,}950 GPU-hours), dominated by
reasoning-mode generation (per-pass ledger: Table~\ref{tab:ledger},
Appendix~\ref{app:repro}). This section shows the cost is
\emph{asymmetric}, and in exactly the direction a practitioner wants.

\subsection{The $k$-sample fidelity curve is asymmetric}
\label{sec:kfidelity}
\begin{figure}[t]
\centering
\includegraphics[width=\linewidth]{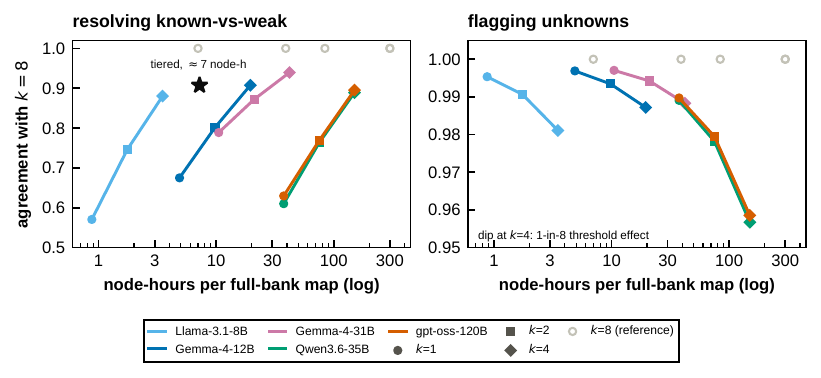}
\caption{Label agreement of random $k$-subsamples (5
repetitions) with the full $k{=}8$ labels, split by reference label
(left: known/weak; right: unknown), against the estimated node-hours of
a full-bank map at that $k$ (Table~\ref{tab:ledger}). Marker shape
encodes $k$ (see legend); open circles are the $k{=}8$ reference, where
agreement is 1 by definition; the star is the tiered $k$-sample budget
of \S\ref{sec:kfidelity} for Gemma-4-12B ($\approx$7 node-h). The right
panel's y-axis spans 0.95--1.00..}
\label{fig:f4}
\end{figure}

We recompute labels from random $k$-subsamples against the $k=\ksamples$
reference. Unconditional $k=1$ agreement looks high (88--96\%) but is
dominated by the unknown mass. Split by reference label, the asymmetry
appears, and it replicates on all five models: on unknowns, $k=1$ agrees
0.989--0.997, while conditioned on the reference being known or weak,
$k=1$ agreement collapses to \konecondrange{}, needing $k=2$ for
\ktwocondrange{} and $k=4$ for \kfourcondrange{} (Figure~\ref{fig:f4}).
Read in GPU-hours, the same fidelity costs the reasoning-mode maps
$\sim$40$\times$ what it costs Llama-3.1.
Recall on a dominant class is cheap, since flagging \emph{everything}
unknown achieves it, so the informative number is precision: $k=1$'s
unknown flags are correct \koneprecrange{} of the time,
+\koneprecliftmin{}--\koneprecliftmax{}pp over the constant-unknown
baseline (\konebaserange), and the lift is largest exactly for the
reasoning models whose unknown mass is smallest.

Independent evidence comes from a fresh-draw retest (\S\ref{sec:limitations}):
unknown labels are \retestunknown{} stable, weak labels only \retestweak.

Finding the boundary is cheap; resolving degrees of knowledge is not. For hedge-tier construction (which facts should this model refuse?), a single sample is enough. Only the small non-unknown slice ($\sim$7--33\%, depending on model and language) needs $k \ge 4$ to separate confident from elicitable knowledge.  A tiered budget ($k=1$ everywhere, escalate the non-unknown slice) cuts labeling cost at matched tier fidelity by an amount that
scales inversely with the escalated slice: \tiersavings{} for the
reasoning-mode maps, whose non-unknown slices are largest, up to
$\approx$5.6$\times$ for Gemma-4-12B ($\approx$7 vs.\ $\approx$39
node-hours, the starred point in Figure~\ref{fig:f4}); replacing the $k=1$ sweep with the
latent probe (\S\ref{sec:latent}) pushes the saving to roughly an order of
magnitude (\S\ref{sec:recipe}).

\subsection{The $k=0$ point: a linear probe on hidden states}
\label{sec:latent}
A per-layer ridge classifier on the pre-generation hidden state separates
known from unknown at peak AUC \latentauclama{} (Llama-3.1-8B, layer
16/32), \latentaucgemma{} (Gemma-4-12B, 33/48), and
\latentaucgemmabig{} (Gemma-4-31B, 39/60), with zero generated
tokens (layer-AUC curves: Figure~\ref{fig:f5}, Appendix~\ref{app:latent}).
Controls rule out surface artifacts: layer-0 and shuffled-label AUC
$\approx$0.5, question-length-only $\le$0.58, popularity-only
$\le$\latentpoponly{}, and length-matched resampling preserves the signal
(31B: \latentlmatchedbig). The probe measures pole discrimination (known
vs.\ $0/\ksamples$ unknown); harder variants (known-vs-rest, three-class)
need weak-tier hidden states we did not extract, and are future work.

Two regularities hold across models. The layer profile is broad and
plateaus past mid-network in all three models (peaks at \latentdepthpct{}
relative depth). And, counterintuitively, peak AUC \emph{declines} with
scale (0.953 $>$ 0.923 $>$ 0.890) across our two families: the larger
model knows more, but what it knows appears less linearly legible from a
single pre-generation state.

\subsection{A practical recipe}
\label{sec:recipe}
Combining both results: (1) run the latent probe (or $k=1$ sampling if
hidden states are unavailable) over the full corpus; (2) treat
high-confidence unknowns as the hedge tier directly; (3) spend $k\ge4$
only where the cheap detector is uncertain or signals knowledge. This
cuts full-corpus mapping from $\approx$39 to $\approx$3 node-hours for a
12B model (latent pass $<$1\,h, then $k{=}4$ over the $\sim$12\% flagged
slice), turning a one-off experiment into routine preprocessing for SFT data 
construction. The estimate assumes the probe's pole-subset fidelity carries to the full corpus; because that fidelity is not directly measured, only the tiered point is drawn in Figure~\ref{fig:f4}, and validating the deployed operating point (weak tier, class imbalance) is future work. One tension is worth stating is that the maps that are most expensive to label, the reasoning-mode MoE models, are the ones the recipe currently helps least; they have the
largest non-unknown slices, and we fit latent probes only on dense
models. Extending the probe to MoE architectures is the highest-value
piece of future work here.
\section{When does reasoning buy recall?}
\label{sec:reasoning}

If long-tail facts are simply absent, reasoning tokens spent on them are
wasted by definition. This section quantifies that waste at full-bank
scale, conditioning every token on the fact's boundary label.

\subsection{Think-on vs.\ think-off at full scale}
\label{sec:thinkdelta}
We probe Gemma-4-12B over the complete bank in both modes (think budget
3{,}000 tokens). Thinking helps, modestly: known mass rises
\gemmaknownar$\to$\thinkknownar{} (Arabic) and
\gemmaknownen$\to$\thinkknownen{} (English), a 38\%/17\% relative gain
(the boundary expansion of \citealp{thinking2recall2026}). But the tail
barely moves: unknown mass stays
at \thinkunkar/\thinkunken{} (ar/en), and the think-on unknown set overlaps
the think-off one at Jaccard \thinkpairunkjacar{} (ar). Thinking also costs
validity: \thinkinvar/\thinkinven{} of records become invalid (truncated
mid-thought; excluded from labels, which \emph{understates} the waste
figures below). Matched-mode substitution shifts cross-family unknown overlaps only slightly and preserves unknown~$>$known in every pair (deltas in
Appendix~\ref{app:maps}), so the structure of \S\ref{sec:boundaries} is
not a mode artifact.

\subsection{Token-waste accounting}
\label{sec:waste}
\begin{figure}[t]
\centering
\includegraphics[width=\linewidth]{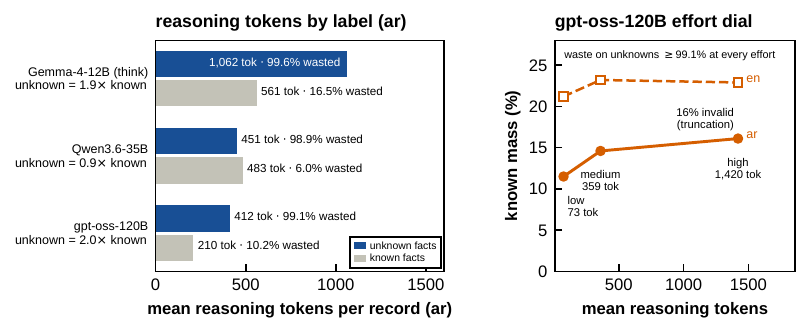}
\caption{Left: mean reasoning tokens per record by boundary label
(Arabic). Gemma-4-12B and gpt-oss-120B think $\sim$2$\times$ harder on
facts they do not know (Qwen3.6 is flat); ``wasted'' = share of those
tokens preceding a wrong answer (unknown-side waste: English within
0.5pp of Arabic). Right: the gpt-oss-120B effort dial (\neffortsubset{} facts): 20$\times$ the tokens
(73$\to$1{,}420) buys $+4.6$pp Arabic known mass, while unknown waste
stays $\ge$\effortwaste{} at every level.}
\label{fig:f6}
\end{figure}
We split every reasoning token by the outcome of the answer it precedes
(Figure~\ref{fig:f6}). The behavioral finding is where models
\emph{choose} to spend: two of three models think hardest exactly where
thinking cannot help.
Gemma-4-12B spends \thinkunkgemma{} mean reasoning tokens on Arabic
unknowns vs.\ \thinkknowngemmatok{} on knowns; gpt-oss-120B spends
\thinkunkgptoss{} vs.\ \thinkknowngptoss{} (Qwen3.6 is flat). Models
escalate compute in response to not knowing; this is BARREL's
``second-thought spiraling'' \citep{yang2025barrel} showing up as token
mass. Nearly all of that escalated spend is lost. Some waste on unknowns
is built into the label definition; what is not built in is its
uniformity across models, languages, and budgets, or the absolute token
masses involved. On unknown facts the share of reasoning tokens
preceding a \emph{wrong} answer is $\ge$\wasteunkmin{} in every model
and language (\wasteunkgemma{} Gemma-4-12B think, \wasteunkqwen{}
Qwen3.6-35B, \wasteunkgptoss{} gpt-oss-120B; Arabic); on known facts
waste drops to \wasteknownrange. Because unknowns dominate the bank,
\wasteoverallqwen--\wasteoverallgemma{} of \emph{all} reasoning tokens
each model generated in this study preceded a wrong answer.

\subsection{The effort dial: marginal recall per reasoning token}
\label{sec:effort}
The reasoning-effort control on gpt-oss-120B lets us trace the
recall-vs-token curve directly. On a stratified \neffortsubset-fact subset
(preserving domain$\times$decile$\times$language composition; medium
reproduces the deploy map's masses to within 0.1pp), raising effort from
low to medium multiplies
mean reasoning length $\sim$5$\times$
(\effortlowthink$\to$\effortmedthink{} tokens) and buys $+3.1$pp Arabic
known mass (\effortlowknownar$\to$\effortmedknownar; English
\effortlowknownen$\to$\effortmedknownen). Low$\to$high multiplies it
$\sim$20$\times$ (\efforthighthink{} mean tokens), yet the medium$\to$high
step adds only $+1.5$pp Arabic (\effortmedknownar$\to$\efforthighknownar)
and \emph{loses} 0.3pp English. Even at an 8{,}000-token budget, high
effort leaves \efforthighinvalidar/\efforthighinvaliden{} (ar/en) of
records invalid and 48.9\% of samples (62.5\% on Arabic unknowns)
truncated in-think (Appendix~\ref{app:probe}), so high-effort recall is
computed on a survivor set and reads optimistic. The
waste rate on unknowns is \effortwaste--\efforthighwaste{} at \emph{every}
effort level (Figure~\ref{fig:f6}, right): the dial changes how long the
model spends before a wrong answer, not whether the answer is wrong.
\citet{zhao2025tts} report the same direction on gpt-oss-20B over SimpleQA
(more thinking $\to$ more hallucination); boundary conditioning at
791K-fact scale adds the mechanism: the marginal token is spent almost
entirely outside the knowledge boundary, where its product is a wrong
answer.

\subsection{A router, simulated from the same samples}
\label{sec:router}

\begin{table}[t]
\centering\footnotesize
\renewcommand{\arraystretch}{0.92}
\begin{tabular}{lrrrrr}
\toprule
System & Acc. & Halluc. & Abstain & R-tok & R-tok/corr.\\
\midrule
\multicolumn{6}{l}{\emph{Gemma-4-12B, full bank (733{,}899 ar records)}}\\
Direct (no-think) & 8.5\% & 90.3\% & 1.1\% & 0 & 0\\
Think-on & 11.9\% & 87.8\% & 0.3\% & 1{,}006 & 8{,}455\\
Gate$\to$answer & 6.9\% & \textbf{1.6\%} & 91.5\% & 0 & 0\\
Gate$\to$think & 6.8\% & 1.7\% & 91.5\% & 57 & 828\\
Gate$\to$retrieve (oracle) & \textbf{87.3\%} & 1.6\% & 11.1\% & 0 & 0\\
\midrule
\multicolumn{6}{l}{\emph{gpt-oss-120B, effort subset (125{,}408 ar records)}}\\
Effort low & 17.9\% & 80.8\% & 1.3\% & 82 & 456\\
Effort medium & 21.6\% & 75.9\% & 2.5\% & 370 & 1{,}716\\
Effort high & 23.8\% & 73.6\% & 2.6\% & 1{,}499 & 6{,}296\\
Gate(low)$\to$medium & 14.3\% & \textbf{3.5\%} & 82.2\% & 126 & 884\\
\bottomrule
\end{tabular}
\caption{Routing simulated from existing probe samples, Arabic (protocol,
assumptions, and English results: Appendix~\ref{app:router}). Row counts
are the records valid in every arm of their block (of \nfacts{} and
\neffortsubset{} ar records respectively; invalid excluded). Gate = one
sampled answer, proceed only if correct; the answer never reuses the gate
sample; reference labels are never consulted. Halluc.\ = confident wrong
answer over \emph{all} queries; R-tok = mean reasoning tokens/query.
Gate$\to$retrieve is an oracle upper bound.}
\label{tab:router}
\end{table}

The samples we already have suffice to simulate the routing they motivate
(Table~\ref{tab:router}). On Gemma-4-12B (ar), gating on a single
no-think sample trades 1.6pp accuracy for a \emph{56$\times$}
hallucination reduction (90.3\%$\to$1.6\%) at zero reasoning tokens; on
gpt-oss, gating at low effort and answering at medium beats effort-high
on every cost column (884 vs.\ 6{,}296 reasoning tokens per correct,
hallucination 3.5\% vs.\ 73.6\%) at 60\% of its accuracy. Thinking
\emph{inside} the gate buys nothing (Gate$\to$think $\approx$
Gate$\to$answer), because the recall thinking buys lies almost entirely
outside the single-sample boundary. Oracle retrieval outside it reaches
87\% accuracy: the map tells a system where retrieval, not reasoning, is
the remedy. For knowledge-shaped queries: hedge or retrieve outside the
boundary, and spend reasoning tokens only inside it.

\section{Discussion}
\label{sec:discussion}

Where the efficient-reasoning literature asks how to think \emph{less} on
problems models can solve, retrieval-shaped queries pose the prior
question of whether to think at all, and \S\ref{sec:reasoning} answers
it: gate on the boundary. The same maps bear on SFT data construction.
\citet{gekhman2024finetuning} showed unknown-fact SFT teaches
hallucination, and AbstentionBench \citep{kirichenko2025abstentionbench}
showed fine-tuning degrades abstention; both argue for
boundary-conditioned data (teach facts inside the boundary, hedges outside
it). Our results make this practical (tiered probing at \tiersavings{}
lower cost) and portable: the transfer test of \S\ref{sec:boundaries}
shows a new model's hedge tier starts at \transferprecrange{} precision
using an existing model's map alone; end-to-end training validation is
future work.

\section{Limitations}
\label{sec:limitations}

\textbf{Grading and provenance.} Alias-augmented substring grading
carries \graderrrange{} population-weighted error against an independent
semantic judge, false-negative-dominant in every map, so unknown masses
are upper bounds within this band; cross-model comparisons of
\emph{absolute} masses inherit up to the between-map error difference
($\le$9pp), while overlap and transfer results use shared fact sets and
are robust to uniform shifts. A shared question/gold/alias bank would
depress all known sets together, inflating unknown-side overlap; the
gold and hedge audits (\S\ref{sec:grading}) bound this channel. The fact
generator (GLM-5.2) contributes fallback gold aliases, but model-free
encyclopedic aliases take precedence, judges are independent of gold
authorship, and the evidence-anchored audit found
\goldauditbad/\goldauditn{} golds contradicted.

\textbf{Label noise.} Every map is a single $k=\ksamples$, $T=\probetemp$
draw, with no per-map error bars. We bound the noise directly: a
5{,}004-fact fresh-draw retest (Gemma-4-12B) gives per-tier stability
\retestunknown/\retestknown/\retestweak{} (unknown/known/weak);
instability concentrates in the weak tier (\S\ref{sec:kfidelity}), and
known-side quantities are read against the retest ceiling throughout.

\textbf{Protocol confounds.} Deploy-default modes and max-token budgets
differ across models; matched-mode checks exist where think passes do
(\S\ref{sec:thinkdelta}), and excluding truncated reasoning
\emph{understates} waste and hides ignorance, not knowledge
(Appendix~\ref{app:maps}). The router of \S\ref{sec:router} is simulated
from probe samples; no deployed system is run.

\textbf{Scope.} One domain (Arab-culture facts from Arabic Wikipedia, a
deliberate cultural-long-tail stress case), two languages, open models
8B--120B (the probe needs weight access); findings may not transfer to
other domains, retrieval settings, or proprietary models.

\section*{Ethics statement}
The fact bank derives from public Arabic Wikipedia content about notable
entities; it contains no private personal data. Boundary maps could in
principle guide targeted misinformation toward what models cannot verify;
we judge the defensive value (hallucination reduction, honest abstention)
to substantially outweigh this. Compute is logged in
Table~\ref{tab:ledger} for replication transparency.

\bibliography{refs}
\bibliographystyle{colm2026_conference}

\appendix
\section{Extended related work}
\label{app:related}

\paragraph{Knowledge boundaries and fine-tuning.}
\citet{gekhman2024finetuning} introduced the sampling-based known/weak/unknown
taxonomy (SliCK) and showed that fine-tuning on unknown facts \emph{teaches}
hallucination; our probe protocol builds on theirs, extended to five models,
two languages, and an alias-robust grader. R-Tuning \citep{zhang2024rtuning}
and refusal-aware variants build abstention SFT from consistency
probes, but for one model, in English, without asking what probing costs.
Boundary-aware training appears in BARREL \citep{yang2025barrel} (SFT+GRPO)
and KARL \citep{gao2026karl} (RL); we stay on the measurement side. See
\citet{li2024kbsurvey} for a survey.

\paragraph{Shared errors vs.\ shared ignorance.}
Recent work shows LLM \emph{errors} correlate across models and providers
\citep{kim2025correlated,goel2025great}, LLMpedia
\citep{saeed2026llmpedia} finds little overlap in what models spontaneously
write about, and concurrent work uses cross-model overlap of rare-fact
knowledge to estimate black-box model capacity \citep{li2026ikp}. We
measure something more specific: boundary-labeled \emph{unknown sets} on a
fixed evidence-grounded fact corpus, where the overlap asymmetry
(unknowns near-identical across families, knowns idiosyncratic) and upward
nesting with scale become visible. That the long tail resists scaling is
well established \citep{kandpal2023longtail,mallen2023popqa,sun2024headtotail};
our decile curves replicate it, and our contribution there is the
\emph{cross-model} structure, not the tail itself.

\paragraph{Cheap boundary detection.}
Pre-generation probes on hidden states are an established alternative to
sampling: \citet{gottesman2024estimating} predict QA success without
generating a token, \citet{chen2026querylevel} detect boundaries from
internal confidence, \citet{kossen2024sep} approximate semantic entropy
with probes, hidden-state signals predict hallucination broadly
\citep{orgad2025llms,gekhman2025insideout}, and boundary probes transfer
across languages \citep{xiao2025crosslingual}. We claim no novelty for
the probe itself. What is missing from this line, and what
\S\ref{sec:cost} contributes, is the \emph{cost/fidelity curve of the
labeling protocol}: how fidelity degrades from $k=\ksamples$ to $k=1$ to
$k=0$ (the probe), its asymmetry between finding unknowns and resolving
known-vs-weak, and the resulting tiered budget.

\paragraph{Self-knowledge and adaptive routing.}
Models partially know what they know
\citep{kadavath2022know,yin2023selfaware}; \citet{cheng2024idk} build
per-model known/unknown maps for ``I don't know'' SFT. Adaptive-RAG
routes between parametric answering and retrieval on self-knowledge or
uncertainty signals \citep{wang2023skr,jeong2024adaptiverag,yao2025seakr},
and \citet{moskvoretskii2025adaptive} find simple uncertainty estimators
the most efficient of 35 such routers. This line supplies the deployment
mechanism our recipe plugs into; we add the boundary-side accounting:
what labels cost at scale, how asymmetrically cheap the unknown side is,
and what reasoning spends when routing is absent.

\paragraph{Reasoning vs.\ knowledge.}
\citet{zhao2025tts} show test-time scaling does not help knowledge-intensive
tasks, including an effort-dial study on gpt-oss-20B where added thinking
converts abstentions into hallucinations; concurrently, the SliCK authors
find thinking expands the parametric recall boundary modestly
\citep{thinking2recall2026}. AbstentionBench
\citep{kirichenko2025abstentionbench} shows reasoning models fail on
unanswerable questions. Our \S\ref{sec:reasoning} instead conditions token
expenditure on \emph{independently measured} boundary labels at 791K-fact
scale, turning BARREL's qualitative failure patterns into token masses.

\paragraph{Efficient reasoning.}
The overthinking literature \citep{sui2025stopoverthinking} taxonomizes
redundant reasoning on problems models \emph{can} solve; certainty-gated
suppression \citep{huang2026cgrs} and capability-boundary monitors
\citep{selfawareness2025} adapt thinking to confidence, mostly on math/code.
For retrieval-shaped queries the prior question is whether to think at all:
when the fact is absent, the efficient budget is zero.

\paragraph{Arabic factuality.}
Existing Arabic hallucination evaluations are small QA/summarization
benchmarks \citep{arahallueval2025}; multilingual local-knowledge benchmarks
\citep{multiloko2025} do not label model-specific boundaries.

\section{Probe and grading details}
\label{app:probe}
\paragraph{Prompts.} Each fact is asked via the model's chat template with a
minimal QA instruction (``Answer the question concisely.'' / Arabic
equivalent), question verbatim, no few-shot examples, no system prompt
beyond the template default.
\paragraph{Reasoning-channel parsing.} Gemma-4: served with
\texttt{skip\_special\_tokens=false}; the \texttt{<|channel>thought}\ldots
\texttt{<channel|>} span is split client-side (default detokenization fuses
thought into the answer). Qwen3.6: split on \texttt{</think>}. gpt-oss:
harmony \texttt{reasoning} field. Llama-3.1: non-reasoning.
\paragraph{Validity and truncation.} A sample is valid iff
\texttt{finish\_reason=stop} or the answer channel is non-empty; records
with $<$50\% valid samples are labeled invalid. Per-map invalid mass:
$<$0.1\% for all no-think maps, 7.2\%/8.0\% (ar/en) for the Gemma think
map at budget 3{,}000, and 30\% (ar) for gpt-oss effort-high at budget
2{,}000, reduced to \efforthighinvalidar{} by the 8{,}000-token re-probe
(\S\ref{sec:effort}). Truncation-in-think rates (sample hits the budget
before leaving the reasoning channel): Gemma think \thinktrunc{} overall
(\thinktruncunk{} on Arabic unknowns); gpt-oss effort low 0.0\%, medium
0.14\%, effort high 48.9\% overall and 62.5\% on Arabic unknowns
even at the widened budget; Qwen3.6's budget-exhausted samples are the
19.7\% discussed below. Probe completeness: 1{,}582{,}119 of 1{,}582{,}120
(fact, language) records per map completed; one English record failed
serving and is excluded everywhere.
\paragraph{Budget-exhausted reasoning (Qwen3.6 only).} When Qwen3.6's
thinking exhausts the token budget before emitting \texttt{</think>}, the
generated text has no channel delimiter and is graded as answer text
(19.7\% of its samples; structurally impossible for gpt-oss and Gemma,
whose channels are separated by the serving stack). Grading over the
reasoning stream is deploy-conservative for unknowns (a gold appearing
mid-reasoning grades correct, so unknown mass is not inflated) and
contributes to Qwen's lower never-right elicitability
(\elicqwenar{} vs.\ 96--98\% for non-reasoning models).
\paragraph{Hedge lexicon.} 13 normalized surface forms (en+ar), hedge
precedence over gold match. The failure mode that could bias our headline
(a hedged-but-correct answer relabeled unknown) is bounded directly by the
exhaustive audit of \S\ref{sec:grading}: $\le$\hedgegoldmax{} of
unknown-labeled facts in the heaviest-hedging cell. Echo-guard: golds
contained in the question (4{,}070 ar / 876 en facts) never match.
\paragraph{Alias generation.} Encyclopedic: arwiki 2026-06 langlinks
(canonical EN title) + redirects (AR surface variants), matched by
normalized title lookup; \naliaswiki{} facts covered; homograph-contested
keys affect 3.1\% of alias-carrying facts (audit shipped with the
release). Model fallback: GLM-5.2 at $T{=}0$, given the gold string,
forbidden from answering; grading uses the union. An earlier Gemma-generated
alias set was retired unused when Gemma became a probed model.

\section{Grader validation detail}
\label{app:grader}
Dual-judge protocol (\S\ref{sec:grading}): 180 samples per (language,
label) cell = 1{,}080 per map. Judge assignment follows two independence
rules: the fact generator (GLM-5.2) never judges, since it authored the
golds, and no judge grades its own family's samples (Qwen3.6-35B judges
the Llama/Gemma/gpt-oss maps; gpt-oss-120B judges Qwen's, plus a shared
slice for inter-judge agreement). The judge prompt asks for semantic equivalence
(translation/transliteration/extra detail = correct) with a one-word
verdict; judge input is the final 2{,}400 characters of the answer channel
(long reasoning answers conclude at the tail). Qwen3.6 judges with thinking
disabled (with it enabled, its preamble echoes the verdict vocabulary and
corrupts parsing); verdict extraction takes the last standalone
CORRECT/INCORRECT token. Table~\ref{tab:graderval} gives per-map results
against the designated judge; errors are population-weighted by the map's
label masses and are false-negative-dominant everywhere.

\begin{table}[h]
\centering\small
\begin{tabular}{llcccc}
\toprule
Map & Judge & Raw agree & Pop.\ FP & Pop.\ FN & Pop.\ error \\
\midrule
Llama-3.1-8B & Qwen3.6-35B & 91.7\% & 0.4\% & 4.3\% & \graderrllama \\
Gemma-4-12B & Qwen3.6-35B & 87.8\% & 1.3\% & 4.5\% & \graderrgemma \\
gpt-oss-120B & Qwen3.6-35B & 86.0\% & 2.8\% & 10.7\% & \graderrgptoss \\
Qwen3.6-35B & gpt-oss-120B & 85.2\% & 3.6\% & 7.2\% & \graderrqwen \\
\bottomrule
\end{tabular}
\caption{Grader validation, $n{=}1{,}080$ per map. FP = grader correct,
judge incorrect; FN = grader incorrect, judge correct (both
population-weighted). On the Llama and Gemma slices judged by \emph{both}
judges, inter-judge agreement is \judgeagreerange{}
($\kappa$ \judgekapparange), and judge choice moves the population-weighted
error by $<$1pp.}
\label{tab:graderval}
\end{table}

\paragraph{Evidence-anchored gold audit.} \goldauditn{} golds sampled
stratified over domain$\times$decile; the judge (Qwen3.6-35B, thinking
disabled) sees the verbatim evidence sentence, the question, and the gold,
and rules SUPPORTED/CONTRADICTED/INSUFFICIENT from the evidence alone.
Result: \goldauditbad{} contradicted (bad-gold rate 0.0\%, rule-of-three
95\% upper bound \goldauditbadub); \goldauditinsuff{} insufficient (the
single extracted sentence does not fully determine the answer, an
extraction-tightness issue rather than a wrong gold). Because the anchor is
Wikipedia text rather than any model's world knowledge, the fact
generator's world model cannot self-confirm here.

\section{Latent probe detail}
\label{app:latent}
\begin{figure}[h]
\centering
\includegraphics[width=0.62\linewidth]{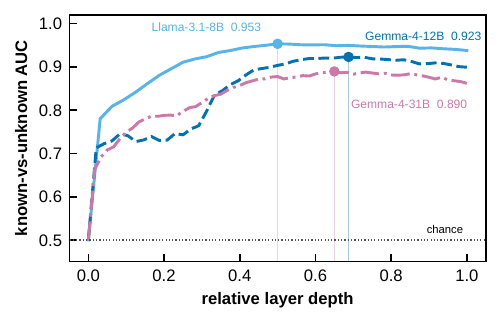}
\caption{Known-vs-unknown AUC of a linear probe by relative layer depth.
Dots and vertical ticks mark each model's peak layer (at
\latentdepthpct{} relative depth). The curves plateau from
mid-network on; peak AUC declines with scale.}
\label{fig:f5}
\end{figure}
Extraction: one forward pass per question (chat template,
\texttt{add\_generation\_prompt}), last-prompt-token hidden state at every
layer, bf16 weights. Classifier: exact ridge regression in Gram space
($\lambda$ swept log-uniformly), 5-fold CV, AUC on held-out folds.
Balanced 5{,}000-fact subsets: 1{,}250 per \{ar,en\}$\times$\{known,
unknown$_{0/8}$\}. Controls per model: layer-0 AUC $\approx$0.5, shuffled
labels $\approx$0.5, constant feature 0.5, question-length-only $\le$0.58,
popularity-only $\le$\latentpoponly{} (0.57/0.60/0.57 for
8B/12B/31B), length-matched resampling \latentlmatchedbig{} (31B). The
probes were fit on labels predating the final alias regrade (masses
shifted $<$0.3\%; the balanced pole subsets are the most label-stable
strata, \S\ref{app:retest}, so the AUCs stand); harder known-vs-rest and
three-class variants require hidden states for the weak tier and are
left to future work.

\section{Label stability}
\label{app:retest}
Seed-retest (Gemma-4-12B, 5{,}004 stratified facts, fresh $k{=}8$ draw at
$T{=}\probetemp$): exact agreement \retestbalanced{} on the stratified
sample (which overweights the rare weak tier by design); per-tier
unknown \retestunknown, known \retestknown, weak \retestweak. Confusion
concentrates in weak$\leftrightarrow$known/unknown; unknown$\to$known flips
are 1 in 1{,}814. Table~\ref{tab:kfid} gives the full $k$-fidelity results
(200K-record subsamples, 5 repetitions). The small dip in unknown-side
agreement at $k=4$ (annotated in Figure~\ref{fig:f4}) is a threshold
artifact: an unknown record with one correct sample in eight flips to
weak exactly when that sample lands in the subsample.
\begin{table}[h]
\centering\small
\begin{tabular}{lccccc}
\toprule
 & \multicolumn{2}{c}{$k{=}1$} & \multicolumn{3}{c}{non-unknown} \\
Model & overall & unknown & $k{=}1$ & $k{=}2$ & $k{=}4$ \\
\midrule
Llama-3.1-8B & 0.945 & 0.995 & 0.57 & 0.75 & 0.88 \\
Gemma-4-12B & 0.959 & 0.997 & 0.67 & 0.80 & 0.91 \\
Gemma-4-31B & 0.959 & 0.997 & 0.79 & 0.87 & 0.94 \\
Qwen3.6-35B & 0.883 & 0.989 & 0.61 & 0.76 & 0.89 \\
gpt-oss-120B & 0.885 & 0.990 & 0.63 & 0.77 & 0.89 \\
\bottomrule
\end{tabular}
\caption{Label agreement of random $k$-subsamples vs.\ the $k{=}8$
reference. Unknown-side agreement is $\ge$0.989 at $k{=}1$ for every model;
resolving the non-unknown tiers needs $k{\ge}4$.}
\label{tab:kfid}
\end{table}

\section{Additional boundary-map detail}
\label{app:maps}
\paragraph{English overlaps.} The pattern of Figure~\ref{fig:f2} replicates
in English: cross-family unknown-Jaccard 0.75--0.87 vs.\ known-Jaccard
0.36--0.61, unknown $>$ known in every pair.
\paragraph{Matched-mode overlaps.} Substituting the Gemma-4-12B think-on
map for its deploy-default map moves cross-family unknown-Jaccard by
0.032/0.038 (ar/en) against Qwen3.6-35B, 0.040/0.043 against gpt-oss-120B,
and 0.096/0.097 against Llama-3.1-8B (thinking prunes Gemma's unknown set
away from the non-reasoning model's). Unknown overlap exceeds known
overlap in every matched-mode pair, so the shared-ignorance result is not
an artifact of comparing reasoning modes, though the Llama pair's raw
magnitude is mode-sensitive.
\paragraph{Hedging rates on unknowns.} Full per-model rates behind
\S\ref{sec:boundaries}: Llama-3.1 hedges on \hedgellamaar{} of Arabic
unknowns vs.\ \hedgellamaen{} of English ones; Qwen3.6 on \hedgeqwenar{}
vs.\ \hedgeqwenen{}; gpt-oss-120B on \hedgegptossar{} vs.\
\hedgegptossen{}; Gemma-4-12B on almost none of either
(\hedgegemmaar/\hedgegemmaen; Gemma-4-31B
\hedgegemmabigar/\hedgegemmabigen).
\paragraph{Think-tail reclaim (R4).} \rfourn{} stratified truncated-invalid
records from the Gemma think map (budget 3{,}000) re-probed at 8{,}000
tokens: \rfourstillinvalid{} still truncate, \rfourtounknown{} resolve to
unknown, \rfourtoknown{} to known ($1$ in $77$). The truncated tail hides
ignorance, not knowledge: excluding it understates unknown mass and token
waste rather than inflating them.

\section{Routing-simulation detail}
\label{app:router}
All quantities are closed-form expectations over hypothetical single
draws from each record's $k$ graded samples ($k_{\text{valid}}$, $n_{\text{correct}}$,
$n_{\text{hedge}}$; hedged samples are never correct by hedge precedence).
Single-draw systems: $P(\text{correct})=n_c/k$,
$P(\text{halluc.})=(k-n_c-n_h)/k$, $P(\text{abstain})=n_h/k$. Same-map
gated systems draw the answer \emph{without replacement}
($P(\text{correct})=\frac{n_c}{k}\cdot\frac{n_c-1}{k-1}$); cross-map
gated systems (gate no-think $\to$ answer think; gate low $\to$ answer
medium) use independent passes
($P(\text{correct})=\frac{n_c^{\text{gate}}}{k^{\text{gate}}}\cdot
\frac{n_c^{\text{ans}}}{k^{\text{ans}}}$). Formulas were verified against
exhaustive enumeration over ordered sample pairs. The $k{=}8$ reference
labels are never used by any router. Reasoning-token costs use the map's
per-(language, label) mean think tokens; the gate's cost is charged on
every query, the answer's only inside the boundary. Records invalid in
any participating map are excluded (7.2\%/8.0\% ar/en for the think pair;
16.4\%/15.7\% for the effort trio, dominated by effort-high truncation),
consistent with the label masses reported elsewhere. Abstention for
ungated systems is the per-\emph{sample} hedge rate (ar 1.2\%), which
differs by definition from the per-\emph{record} hedge rates of
\S\ref{sec:boundaries} (a record hedges if any of its 8 samples do).
Gate$\to$retrieve assumes retrieval returns the fact's stored evidence
span and the model answers correctly whenever the span suffices
(\goldsuffrate{} of audited golds), abstaining otherwise; this is an
oracle upper bound, not a retrieval system. English results
(Table~\ref{tab:routeren}) reproduce the Arabic ordering, with a
63$\times$ hallucination reduction for the Gemma gate.

\begin{table}[h]
\centering\footnotesize
\begin{tabular}{lrrrrr}
\toprule
System & Acc. & Halluc. & Abstain & R-tok & R-tok/corr.\\
\midrule
\multicolumn{6}{l}{\emph{Gemma-4-12B, full bank (727{,}373 en records)}}\\
Direct (no-think) & 12.0\% & 88.0\% & 0.0\% & 0 & 0\\
Think-on & 15.0\% & 84.8\% & 0.2\% & 844 & 5{,}631\\
Gate$\to$answer & 10.6\% & 1.4\% & 88.0\% & 0 & 0\\
Gate$\to$think & 10.3\% & 1.7\% & 88.0\% & 60 & 585\\
Gate$\to$retrieve (oracle) & 88.0\% & 1.4\% & 10.6\% & 0 & 0\\
\midrule
\multicolumn{6}{l}{\emph{gpt-oss-120B, effort subset (126{,}392 en records)}}\\
Effort low & 28.6\% & 70.1\% & 1.3\% & 63 & 221\\
Effort medium & 31.2\% & 66.9\% & 1.9\% & 332 & 1{,}064\\
Effort high & 31.4\% & 66.3\% & 2.3\% & 1{,}342 & 4{,}271\\
Gate(low)$\to$medium & 24.9\% & 3.6\% & 71.5\% & 121 & 485\\
\bottomrule
\end{tabular}
\caption{Routing simulation, English (protocol above; cf.\
Table~\ref{tab:router}).}
\label{tab:routeren}
\end{table}

\section{Reproducibility and compute}
\label{app:repro}
\paragraph{Models.} Table~\ref{tab:models} lists the exact model variants
used. All probed models are open-weight instruction-tuned checkpoints
obtained from Hugging Face; GLM-5.2 generates facts and fallback aliases
only and is never probed or used as a judge.

\begin{table}[h]
\centering\footnotesize
\renewcommand{\arraystretch}{1.0}
\begin{tabular}{llll}
\toprule
Name in paper & Hugging Face repository & Role & Reference \\
\midrule
Llama-3.1-8B & \texttt{meta-llama/Llama-3.1-8B-Instruct} & probed & \citet{llama3herd2024} \\
Gemma-4-12B & \texttt{google/gemma-4-12B-it} & probed & \citet{gemma4report2026} \\
Gemma-4-31B & \texttt{google/gemma-4-31B-it} & probed & \citet{gemma4report2026} \\
Qwen3.6-35B & \texttt{Qwen/Qwen3.6-35B-A3B} & probed, judge & \citet{qwen36blog2026} \\
gpt-oss-120B & \texttt{openai/gpt-oss-120b} (mxfp4) & probed, judge & \citet{gptoss2025} \\
GLM-5.2 & \texttt{zai-org/GLM-5.2} & generator only & \citet{glm5report2026} \\
\bottomrule
\end{tabular}
\caption{Exact model variants. ``Judge'' models grade only maps outside
their own family (\S\ref{sec:grading}); the generator is excluded from
all evaluation. Exact snapshot hashes ship with the code release.}
\label{tab:models}
\end{table}

\paragraph{Serving.} vLLM \citep{kwon2023efficient} (OpenAI-compatible),
one fleet per model:
Llama-3.1-8B-Instruct TP1$\times$8/node; Gemma-4-12B-it TP1$\times$8;
Gemma-4-31B-it TP2$\times$4; Qwen3.6-35B TP4$\times$2; gpt-oss-120B
(mxfp4) TP4$\times$2 or TP2$\times$4, 16K context. Exact snapshot hashes
and launch flags ship with the code release. Sampling: $k{=}8$,
$T{=}\probetemp$, top-$p$ 0.95, max-tokens 1{,}200 (no-think) / 2{,}000
(gpt-oss) / 3{,}000 (Gemma think) / 8{,}000 (truncation re-probes).
\paragraph{Compute.} All experiments ran on 8$\times$H100 nodes. Total
serving compute for every result in this paper: \textbf{1{,}243 node-hours
($\approx$9{,}950 H100-hours)}, measured from cluster job accounting;
Table~\ref{tab:ledger} itemizes per-pass costs at observed
throughput.

\begin{table}[h]
\centering\footnotesize
\renewcommand{\arraystretch}{0.92}
\begin{tabular}{lcc}
\toprule
Pass (1.58M records unless noted) & node-h & driver \\
\midrule
Llama-3.1-8B ($k{=}8$) & $\approx$7 & non-reasoning, 1{,}200 tok \\
Gemma-4-12B no-think ($k{=}8$) & $\approx$39 & 1{,}200 tok \\
Gemma-4-12B think ($k{=}8$) & $\approx$140 & 3{,}000 tok \\
Gemma-4-31B ($k{=}8$) & $\approx$84 & 1{,}200 tok \\
Qwen3.6-35B ($k{=}8$, reasoning) & $\approx$250--350 & 3{,}000 tok \\
gpt-oss-120B ($k{=}8$, reasoning) & $\approx$250--350 & 2{,}000 tok \\
GLM-5.2 alias generation & $\approx$140 & one-off, non-probed \\
Effort trio, re-probes, latent, validation & $\approx$130 & \S\ref{sec:effort}, App.~\ref{app:maps} \\
\midrule
Total, all experiments (job accounting) & \textbf{1{,}243} & $\approx$9{,}950 H100-hours \\
\bottomrule
\end{tabular}
\caption{GPU-hour ledger (H100 node-hours; 1 node = 8 GPUs). Rows are
throughput-derived estimates; the total is from job accounting.
Reasoning-mode generation dominates; the routing of \S\ref{sec:recipe}
avoids exactly the expensive rows.}
\label{tab:ledger}
\end{table} The tiered budget of \S\ref{sec:recipe} scales these by the
sample-count reduction; the latent probe costs one forward pass per fact
($<$1 node-h per model at batch inference) plus a one-off labeled subset.
\paragraph{Release.} Fact bank (questions, golds, evidence spans,
popularity, domain), all six boundary maps, alias sets with per-source
provenance, grader/probe/analysis code, and per-figure data.

\section{Licensing}
\label{app:ethics}
Fact content derives from Arabic Wikipedia (CC BY-SA 4.0); we release
derived questions/golds under the same license with attribution. Generated
questions were produced by GLM-5.2 consistent with its license terms.

\section{LLM usage disclosure}
\label{app:llmuse}
LLMs are part of the measurement pipeline itself: an open model generates
the fact bank and gold aliases, and two others serve as grading judges,
as described in \S\ref{sec:methods}. Beyond that, we used LLM coding
assistants to help implement the experiments and analysis code, and an
LLM to help polish the writing of this paper. All experimental design,
results, and claims were authored and verified by the authors.

\end{document}